\pdfoutput=1
\RequirePackage{amsmath}         %

\documentclass[runningheads,a4paper]{llncs}

\usepackage{graphicx}            %
\usepackage{amssymb}             %
\usepackage{amsfonts}            %
\usepackage{enumitem}            %
\usepackage{multirow}            %
\usepackage{siunitx}             %
\usepackage{url}                 %
\usepackage{xspace}              %
\usepackage[T1]{fontenc}         %
\usepackage[hidelinks]{hyperref} %
\usepackage{wrapfig}

\graphicspath{{images/}}
\DeclareGraphicsExtensions{.pdf,.png,.jpg,.jpeg}

\newcommand{\degreem}{^{\circ}} %

\newcommand{\seclabel}[1]{\label{sec:#1}}

\newcommand{\figlabel}[1]{\label{fig:#1}}

\newcommand{\secref}[1]{Section~\ref{sec:#1}\xspace}

\newcommand{\figref}[1]{Fig.~\ref{fig:#1}\xspace}

\newcommand{\nop}{NimbRo\protect\nobreakdash-OP\xspace}

\newcommand{\cm}{CM730\xspace}
\newcommand{\cmnew}{CM740\xspace}

\newcommand{\igus}{igus\textsuperscript{\tiny\circledR}\xspace}
\newcommand{\iguhop}{igus\textsuperscript{\tiny\circledR}$\!$ Humanoid Open Platform\xspace}

\newcommand{\degree}{$\degreem$\xspace}

\usepackage{eso-pic}

\AtBeginDocument{\AddToShipoutPictureFG*{\AtTextUpperLeft{\put(0,\LenToUnit{40pt}){\parbox{\textwidth}{\centering\bfseries
RoboCup 2016: Robot World Cup XX, Lecture Notes in\\Computer Science 9776, Springer, 2017
}}}}}
\begin{document}

\mainmatter

\title{RoboCup 2016 Humanoid TeenSize Winner NimbRo: Robust Visual Perception and Soccer Behaviors}
\titlerunning{RoboCup 2016 TeenSize Winner NimbRo}

\author{Hafez Farazi, Philipp Allgeuer, Grzegorz Ficht, Andr\'{e} Brandenburger,\\Dmytro Pavlichenko, Michael Schreiber and Sven Behnke}
\authorrunning{Farazi, Allgeuer, Ficht, Brandenburger, Pavlichenko, Schreiber, Behnke}

\institute{Autonomous Intelligent Systems, Computer Science, Univ.\ of Bonn, Germany\\
\url{{farazi, pallgeuer}@ais.uni-bonn.de}, \url{behnke@cs.uni-bonn.de}\\
\url{http://ais.uni-bonn.de}}

\maketitle

\begin{abstract}
The trend in the RoboCup Humanoid League rules over the past few years has been 
towards a more realistic and challenging game environment. Elementary skills 
such as visual perception and walking, which had become mature enough for 
exciting gameplay, are now once again core challenges. The field goals are both white,
and the walking surface is artificial grass, which constitutes 
a much more irregular surface than the carpet used before. In this 
paper, team NimbRo TeenSize, the winner of the TeenSize class of the RoboCup 
2016 Humanoid League, presents its robotic platforms, the adaptations that had 
to be made to them, and the newest developments in visual perception and soccer behaviour.
\end{abstract}

\section{Introduction}

\begin{figure}[!b]
\parbox{\linewidth}{\centering
\includegraphics[height=44mm]{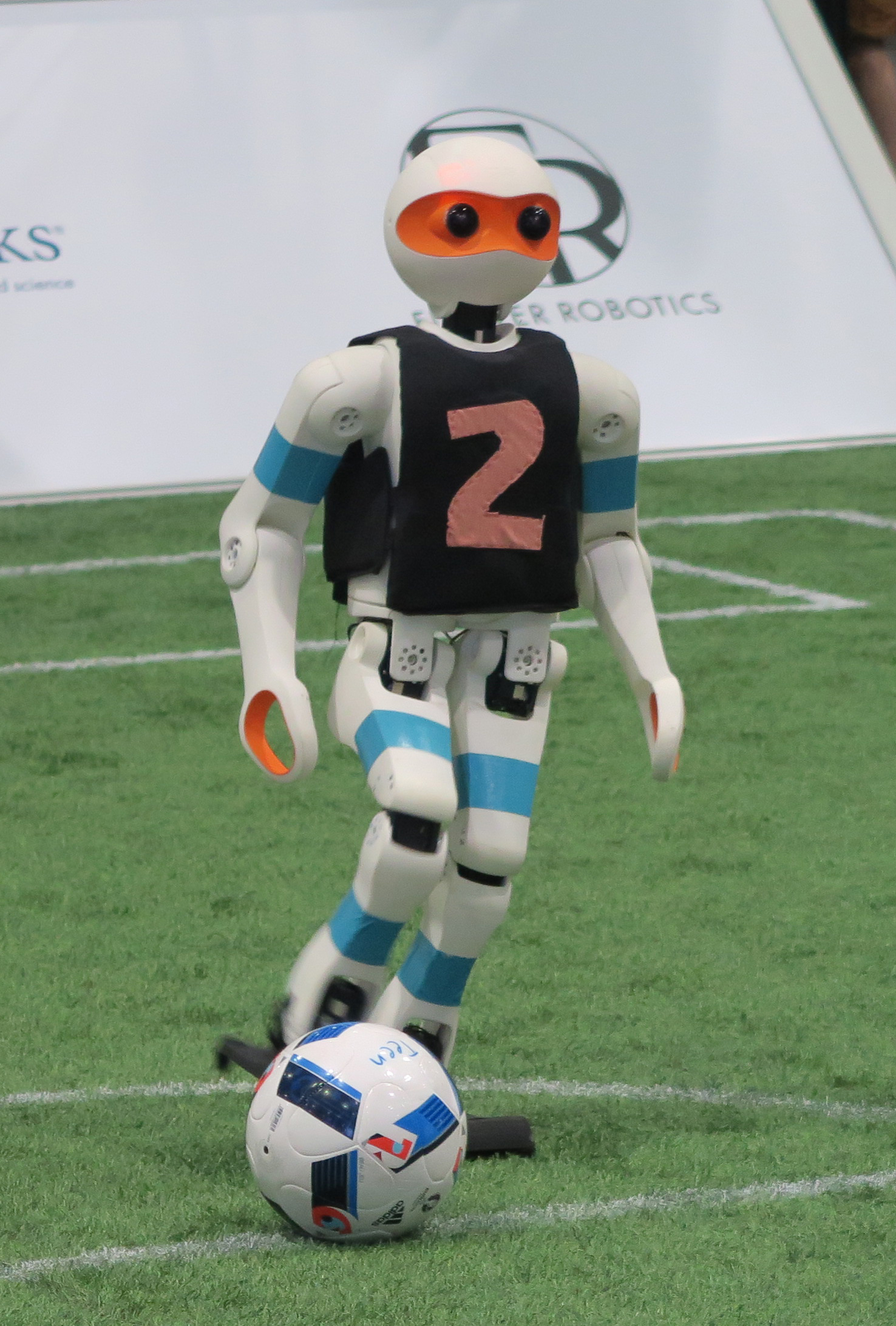}\hspace{1.3mm}%
\includegraphics[height=44mm]{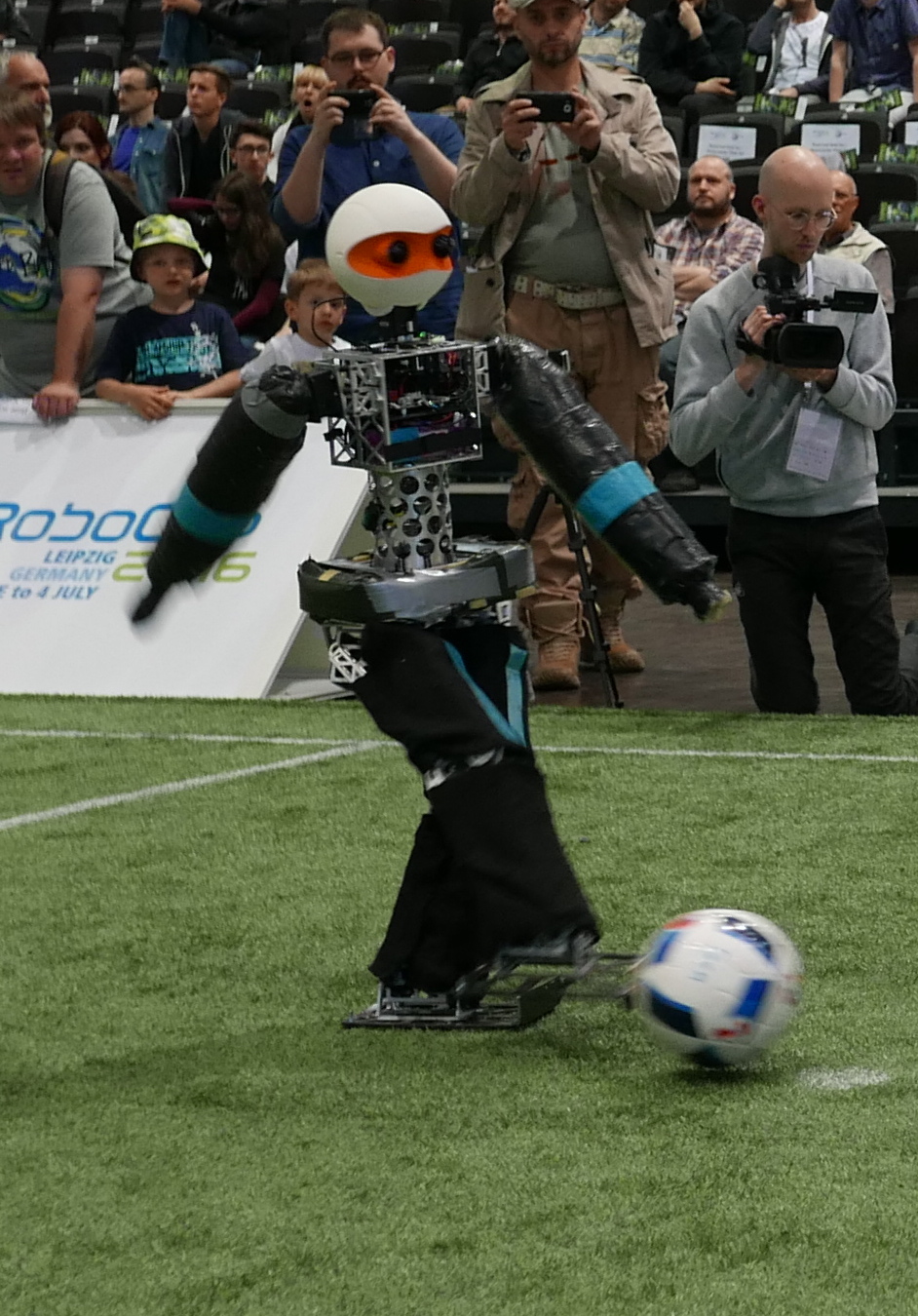}\hspace{1.3mm}%
\includegraphics[height=44mm]{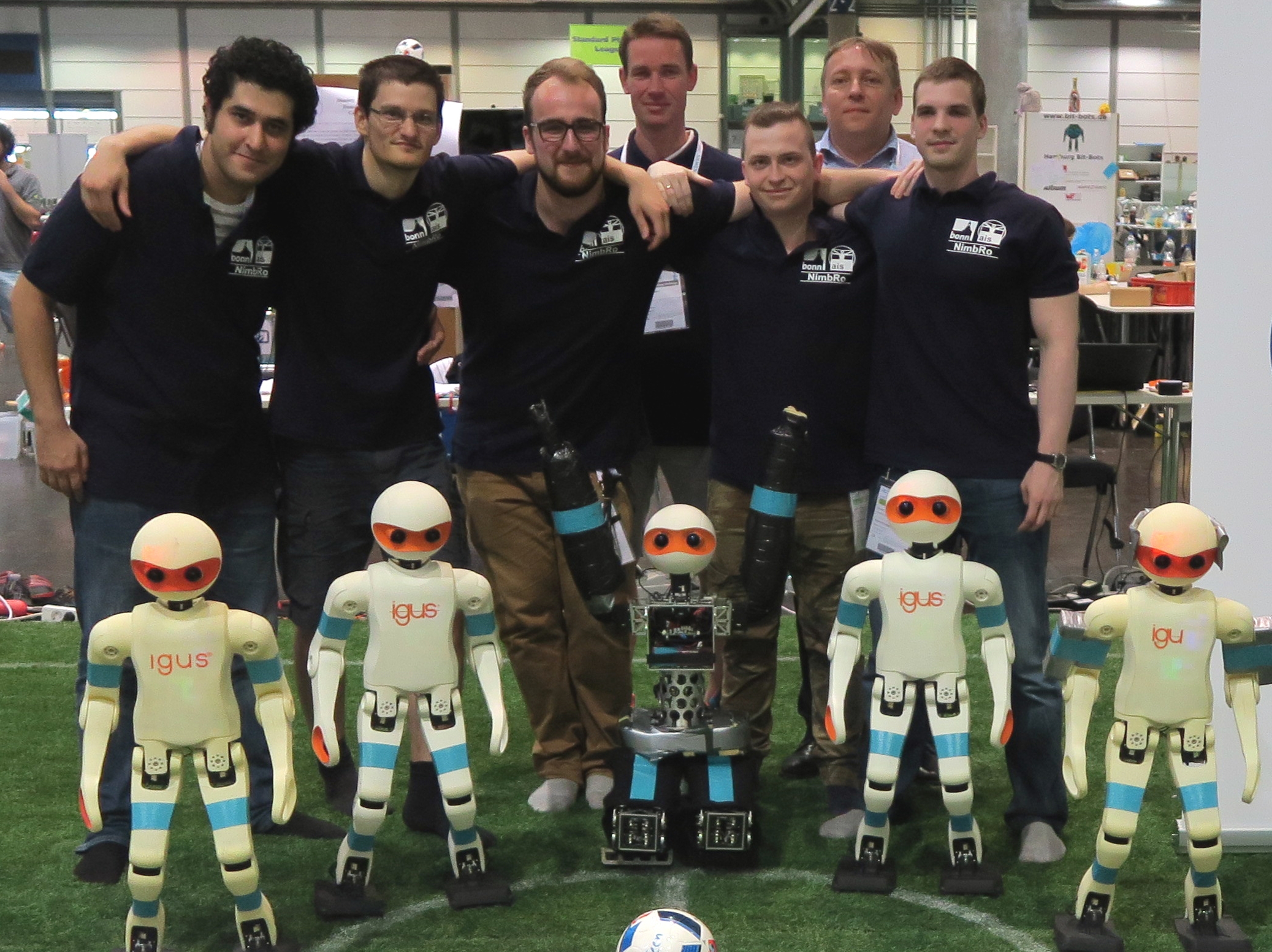}}
\caption{The \iguhop, Dynaped and team NimbRo.}
\figlabel{robots_team}
\end{figure}

In the RoboCup Humanoid League, there is an ongoing effort to develop humanoid 
robots capable of playing soccer. There are three size classes---KidSize 
($40$-$90$cm), TeenSize ($80$-$140$cm), and AdultSize ($130$-$180$cm). Note that 
the overlap in size classes is intentional to facilitate teams to move up to higher size classes.

The TeenSize class of robots started playing 2 vs.\ 2 games in 2010, and a year 
later moved to a larger soccer field. In 2015, numerous changes were made in the 
rules that affected mainly visual perception and walking, namely:
\begin{itemize}
\item the green field carpet was replaced with artificial grass,
\item the field lines are only painted on the artificial grass, so they are 
highly variable in appearance and no longer a clear white,
\item the only specification of the ball is that it is at least 50\% white (no 
longer orange), and
\item the goal posts on both sides of the field are white.
\end{itemize}
Our approach to addressing these rule changes in terms of perception are given 
in \secref{perception}.

In this year's RoboCup, we used our fully open-source 3D printed robots, the 
\iguhop \cite{Allgeuer2015b}, and the associated open-source ROS software. 
Furthermore, we revived one of our classic robots, Dynaped, with 
the same ROS software, by upgrading its electronics and PC and developing a new 
communications scheme for this to work. This was done in such a way that the 
robot hardware in use was completely transparent to the software, and abstracted 
away through a hardware interface layer. More details are given in 
\secref{robot_platforms}. Both platforms are shown in 
\figref{robots_team}, along with the human team members.

\section{Robot Platforms}
\seclabel{robot_platforms}

\subsection{Igus Humanoid Open Platform}
\seclabel{igus}

RoboCup 2016 was the first proper debut of the latest addition to the NimbRo 
robot soccer family---the \iguhop, shown on the left in \figref{robots_team}. 
Although an earlier version of the robot had briefly played in 2015, 2016 was 
the first year where the platform constituted an integral part of the NimbRo 
soccer team. Over the last three years, the platform has seen continual 
development, originating from the initial \nop prototype, and seven robots of three generations have 
been constructed. The \iguhop is \SI{92}{\centi\meter} tall and weighs only 
\SI{6.6}{\kilogram} thanks to its 3D printed plastic exoskeleton design. The 
platform incorporates an Intel Core i7-5500U CPU running a full 64-bit Ubuntu 
OS, and a Robotis \cm microcontroller board, which electrically interfaces the 
twelve MX-106R and MX-64R RS485 servos. The \cm incorporates 3-axis 
accelerometer, gyroscope and magnetometer sensors, for a total of 9 axes of 
inertial measurement. For visual perception, the robot is equipped with a 
Logitech C905 USB camera fitted with a wide-angle lens. The robot software is 
based on the ROS middleware, and is a continuous evolution of the ROS software 
that was written for the \nop. The \iguhop is discussed in greater detail in 
\cite{Allgeuer2015b}.

\subsection{Upgraded Dynaped}
\seclabel{dynaped}

Dynaped, shown in the middle in \figref{robots_team}, has been an active player 
for team NimbRo since RoboCup 2009 in Graz, Austria. Through the years, Dynaped 
has played both as a goalie and a field player during numerous competitions, 
contributing to the team's many successes. Dynaped's competition performance and 
hardware design, including features like the effective use of parallel 
kinematics and compliant design, contributed to NimbRo winning the Louis Vuitton Best Humanoid Award 
in both 2011 and 2013.

In 2012, our focus of development shifted towards the development of an open 
platform---the \nop prototype, later followed by the \iguhop. Because of 
platform incompatibilities and a severe electrical hardware failure 
during RoboCup 2015, we decided to upgrade Dynaped to the newer ROS software, in a way that is 
transparently compatible with the \iguhop. The upgrade 
included both hardware and software improvements. In terms of hardware, Dynaped has 
been equipped with a modern PC (Intel Core i7-5500U CPU), a new \cmnew 
controller board from Robotis, and a vision system as in the \iguhop, consisting 
of the same camera unit and 3D-printed head.

\begin{figure}[!tb]
\centering
\includegraphics[height=42mm]{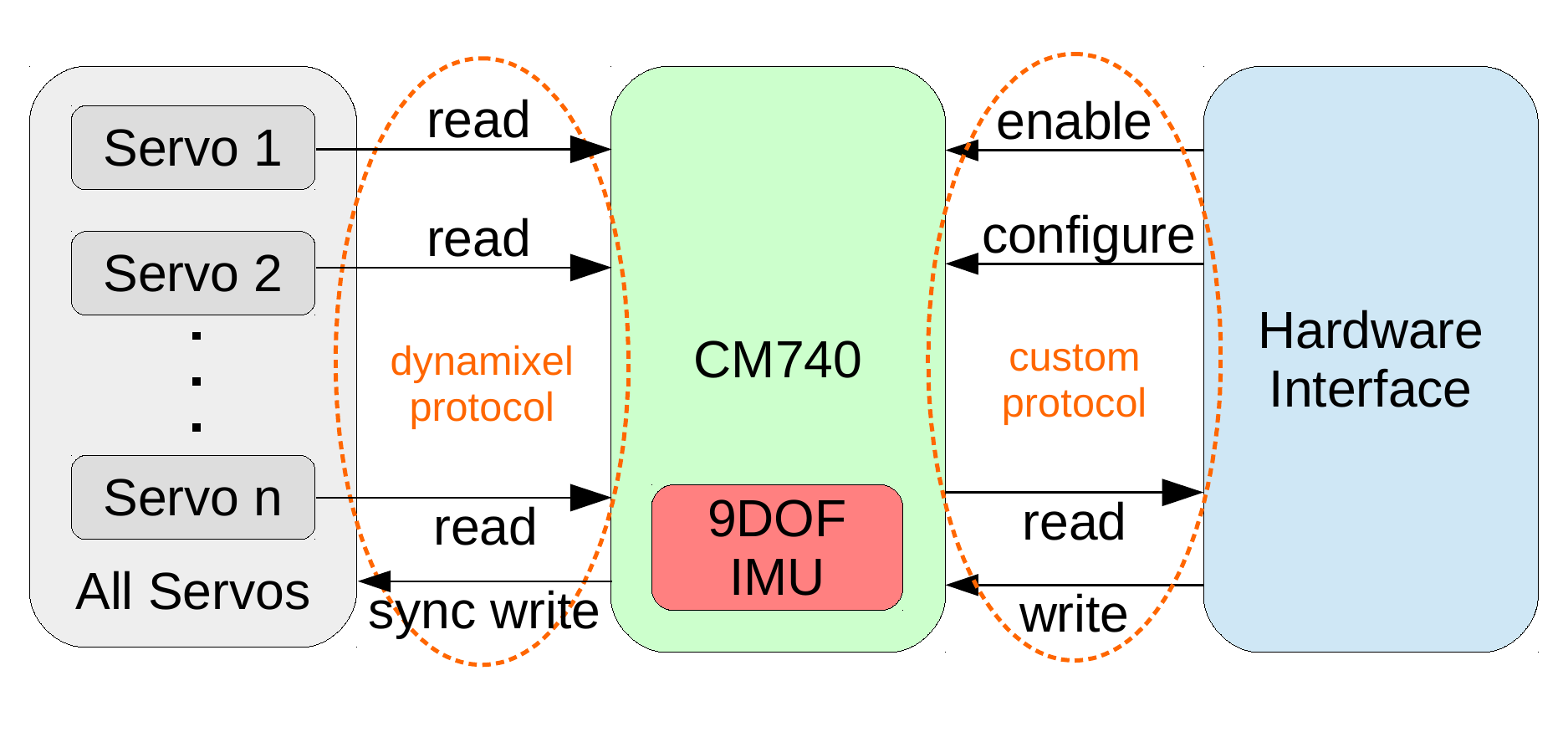}\hspace{1.5mm}%
\vspace{-12pt}
\caption{Dynaped's custom communication scheme}
\figlabel{megapacket}
\vspace{-10pt}
\end{figure}

To adapt Dynaped to the new ROS software framework, a number of modifications 
had to be made. Thanks to the modularity of our software, only the low-level 
modules needed a specific reimplementation, while all of our high-level 
functionality that contributed to the success during RoboCup 2016 could be used 
untouched. At first glance, the only fundamental difference seems to be the 
utilisation of parallel kinematics, leading to the loss of one degree of 
freedom, but in fact, quite importantly, Dynaped still uses the older Dynamixel 
actuators. The used Dynamixel EX-106 and RX-64 have very similar physical 
properties to the MX-106 and MX-64 used in the \iguhop, but they lack a bulk 
read instruction, which is essential for allowing fast communications with 
multiple actuators with a single instruction. This limitation greatly reduces 
the control loop frequency, as each actuator needs to be read individually. This 
increases the latencies with each added actuator to the bus. To reduce these 
delays, Dynaped utilises a custom firmware for the \cmnew, which no longer acts 
merely as a passthrough from the PC to the servos. Instead, it communicates with 
both sides in parallel (see \figref{megapacket}). On the actuator side, the 
\cmnew queries all registered devices on the Dynamixel bus in a loop. 
Communications with the PC are performed using an extension of the original 
Dynamixel protocol, which allows the use of the same, well-developed error 
handling as our original firmware, as well as having the option to still use the 
CM740 as a passthrough device. The \cmnew-PC protocol has been extended by four 
new instructions: 

\begin{itemize}
\item Configure extended packet communication,
\item Enable extended packet communication,
\item Send extended packet, and
\item Receive extended packet.
\end{itemize}

To start the custom communication scheme, the hardware interface first sends a 
configuration packet containing a list of servo ID numbers, along with their 
respective model types. This informs the \cmnew which servo registers it needs 
to keep reading from and writing to. Typically, these registers correspond to 
position, torque, and controller gain data. The read packets contain the most recent data from all of 
the Dynamixel devices, with an indication of how many times it has been read 
from since the last packet. The write packets include the current position 
setpoints and compliance values for the servos. In Dynaped's case, the packet 
transmission frequency is 100Hz, which allows all devices on the Dynamixel bus 
to be read at least once before a new read packet is sent. This transfer rate 
would not be achievable on Dynaped's hardware with the traditional 
request-response transmission paradigm.

\begin{figure}[!tb]
\centering
\vspace{-10pt}
\includegraphics[height=42mm]{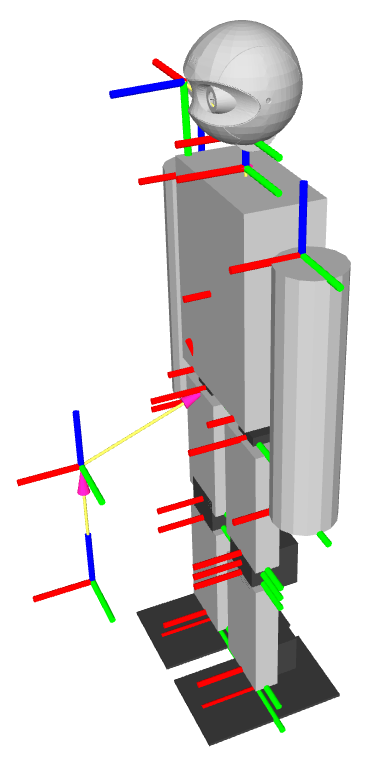}\hspace{1.5mm}%
\includegraphics[height=42mm]{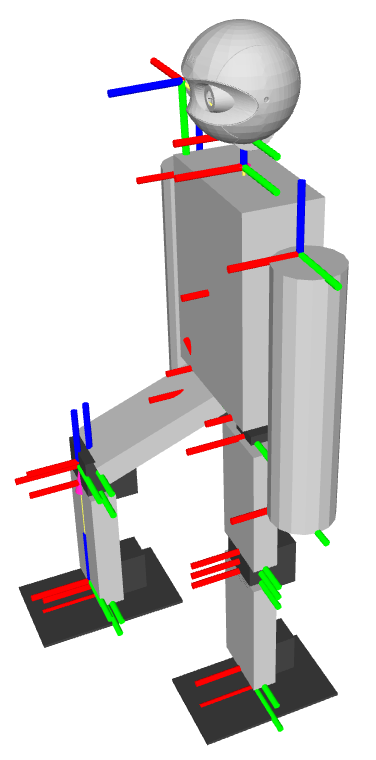}%
\vspace{-12pt}
\caption{Serial to parallel kinematic translation. Left: Before translation, virtual chain visible. Right: After translation.}
\figlabel{dynaped_kinematics}
\vspace{-12pt}
\end{figure}

Creating a model with parallel kinematics and using it in the hardware interface 
proved to be another challenge, as ROS does not natively support this. In order 
to translate between the serial and parallel kinematics, the created model has 
two sets of leg kinematic chains, a virtual serial one and the true parallel one 
that is actuated. The virtual kinematic chain receives the commands as-is from 
the motion modules, which the hardware interface then translates for the 
parallel chain (see \figref{dynaped_kinematics}) before sending a command to the 
actuators. In order to recreate the state of the parallel joints when reading 
out positions from the actuators, virtual joints have been added in order to 
offset the next link in the kinematic chain by the same angle that the joint has 
rotated. With these modifications, the robot can be seen as an \iguhop robot by 
the software, and thanks to our modular design approach, no robot-specific 
changes had to be done to any motion modules, or other higher level parts of our 
code.

\section{Software Design}
\seclabel{software_design}

\subsection{Visual Perception}
\seclabel{perception}

The primary source of perceptual information for humanoid robots on the soccer field
is the camera. Each robot is equipped with one Logitech C905 camera, fitted 
with a wide-angle lens that has an infrared cut-off filter. The diagonal field of view is approximately 150\degree. The choice of lens was optimised 
to maximise the number of usable pixels and minimise
the level of distortion, without significantly sacrificing the effective 
field of view. Our vision system is able to detect the field boundary, line 
segments, goal posts, QR codes and other robots using texture, shape, brightness 
and colour information. After identifying each object of interest, by using appropriate 
intrinsic and extrinsic camera parameters, we project each object into egocentric world 
coordinates. The intrinsic camera parameters are pre-calibrated, but the extrinsic 
parameters are calculated online by consideration of the known kinematics and estimated orientation of the robot.
Although we have the kinematic model of both robot platforms, 
some variations still occur on the real hardware, resulting in
projection errors, especially for distant objects. To address this, we utilised the Nelder-Mead \cite{nelder1965simplex} method to calibrate the position and orientation 
of the camera frame in the head. This calibration is crucial for 
good performance of the projection operation from pixel coordinates to 
egocentric world coordinates, as demonstrated in \figref{kin_calib}.
As a reference, the raw captured image used to 
generate the figure is shown in the left side of \figref{vision_output}.
More details can be found in \cite{farazi2015}.

\begin{figure}[!ht]
\parbox{\linewidth}{\centering
\includegraphics[height=4.65cm]{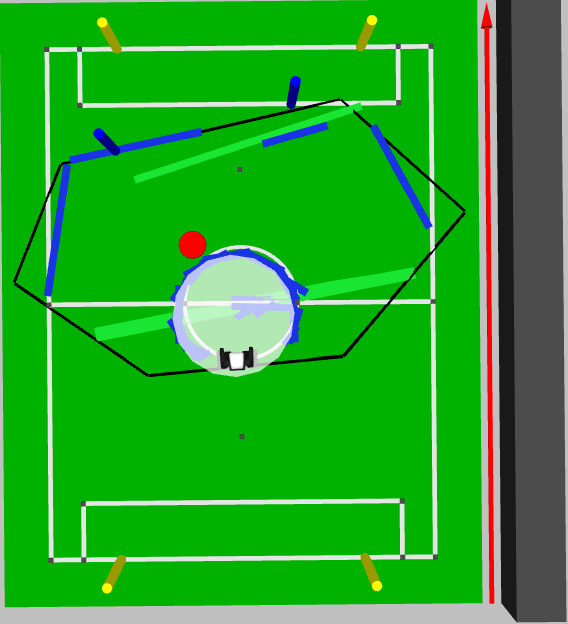}\hspace{0.02\linewidth}\includegraphics[height=4.65cm]{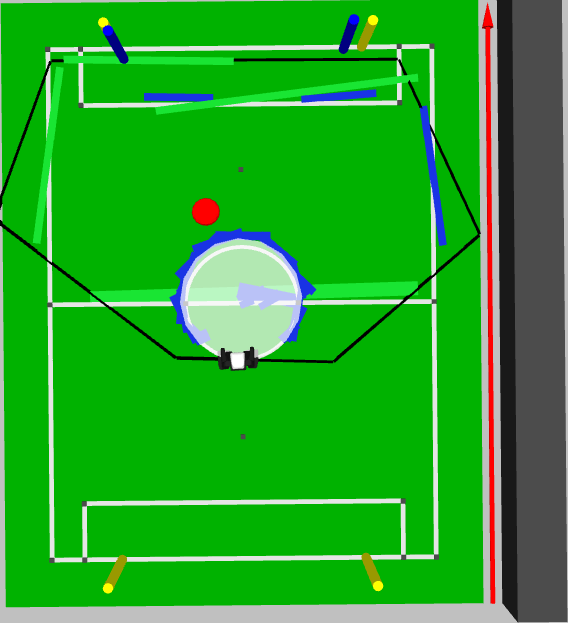}}
\caption{Projected ball, field line and goal post detections before (left) and after (right) kinematic calibration.}
\figlabel{kin_calib}
\vspace{-7ex}
\end{figure}
\begin{figure}[!t]
\parbox{\linewidth}{\centering
\includegraphics[width=0.49\linewidth]{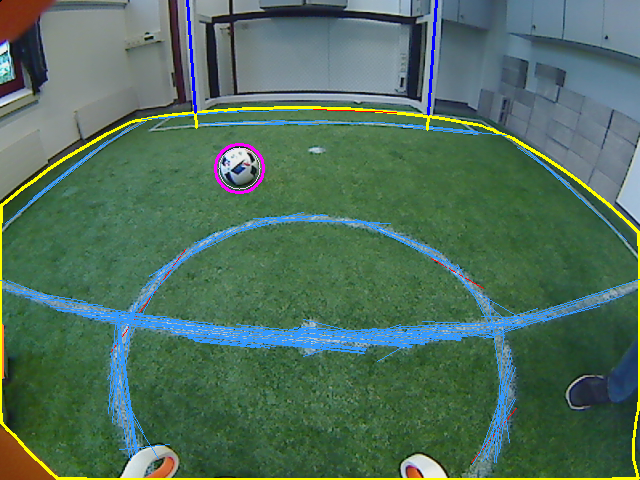}\hspace{0.019\linewidth}\includegraphics[width=0.49\linewidth]{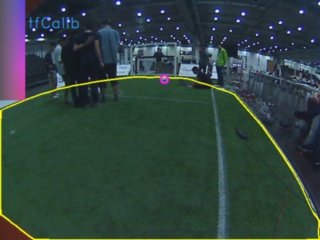}}
\caption{Left: A captured image with ball (pink circle), field line (light blue lines), field boundary (yellow lines), and goal post (dark blue lines) detections.
Right: Distant ball detection on RoboCup 2016 field.}
\figlabel{vision_output}
\vspace{-2ex}
\end{figure}

\subsubsection{Field Detection:}

Although it is a common approach for field boundary detection to find the convex 
hull of all green areas directly in the image \cite{laue2009efficient}, more 
care needs to be taken in our case due to the significant image distortion. The 
convex hull may include parts of the image that are not the field. To exclude 
these unwanted areas, vertices of the connected regions are first undistorted 
before calculating the convex hull. The convex hull points and intermediate 
points on each edge are then distorted back into the raw captured image, and the 
resulting polygon is taken as the field boundary. An example of the final 
detected field polygon is shown in \figref{vision_output}.

\subsubsection{Ball Detection:}

In previous years of the RoboCup, most teams used simple colour segmentation and 
blob detection-based approaches to find the orange ball. 
Now that the ball has a pattern and is mostly white however, such simple 
approaches no longer work effectively, especially since the lines and goal posts 
are also white. We extend \cite{schulz2007ball}, our approach is divided into two stages. In the first stage, 
ball candidates are generated based on colour segmentation, colour histograms, 
shape and size. White connected components in the image are found, and the 
Ramer-Douglas-Peucker \cite{ramer1972iterative} algorithm is applied to reduce the number of polygon 
vertices in the resulting regions. This is advantageous for quicker subsequent 
detection of circle shapes. The detected white regions are searched for at least 
one third full circle shapes within the expected radius ranges. Colour 
histograms of the detected circles are calculated for each of the three HSV 
channels, and compared to expected ball colour histograms using the 
Bhattacharyya distance. Circles with a suitably similar colour distribution to the
expected one are considered to be ball candidates.

In the second stage of processing, a dense histogram of oriented gradients (HOG) 
descriptor \cite{dalal2006object} is applied in the form of a cascade 
classifier, with use of the AdaBoost technique. Using this cascade classifier, 
we reject those candidates that do not have the required set of HOG features. The aim of using the HOG 
descriptor is to find a description of the ball that is largely invariant to 
changes in illumination and lighting conditions. The HOG descriptor is not 
rotation invariant, however, so to detect the ball from all angles, and to 
minimise the user's effort in collecting training examples, each positive image 
is rotated by $\pm$10\degree and $\pm$20\degree and mirrored horizontally, with 
the resulting images being presented as new positive samples, as shown in \figref{ball_samples}. Greater rotations 
are not considered to allow the cascade classifier to learn the shadow under the 
ball. The described approach can detect balls with very few false positives, 
even in environments cluttered with white, and under varying lighting conditions. 
In our experiments, we observed detection a FIFA size 3 ball up 
to \SI{4.5}{\metre} away with a success rate above 80\% on a walking robot, and up to \SI{7}{\metre} away on a stationary robot, as shown in the right side of \figref{vision_output}. It is 
interesting to note that our approach can find the ball in undistorted and 
distorted images with the same classifier.

\begin{figure}[!tb]
\parbox{\linewidth}{\centering
\includegraphics[width=0.088\linewidth]{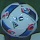}
\includegraphics[width=0.088\linewidth]{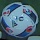}
\includegraphics[width=0.088\linewidth]{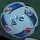}
\includegraphics[width=0.088\linewidth]{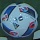}
\includegraphics[width=0.088\linewidth]{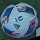}
\includegraphics[width=0.088\linewidth]{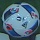}
\includegraphics[width=0.088\linewidth]{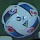}
\includegraphics[width=0.088\linewidth]{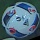}
\includegraphics[width=0.088\linewidth]{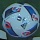}
\includegraphics[width=0.088\linewidth]{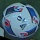}}
\caption{Ball detection tracking data augmentation extending one positive sample (leftmost) to ten, by applying rotations and mirroring operations.}
\figlabel{ball_samples}
\vspace{-0.8em}
\end{figure}
\subsubsection{Field Line and Centre Circle Detection:}

Due to the introduction of artificial grass in the RoboCup Humanoid League, the 
field lines are no longer clearly visible. In past 
years, many teams based their line detection approaches on the segmentation of 
the colour white \cite{laue2009efficient}. This is no longer a robust approach 
due to the increased number of white objects on the field, and due to the visual 
variability of the lines. Our approach is to detect spatial changes in 
brightness in the image using a Canny edge detector on the V channel of the HSV 
colour space. The V channel encodes brightness information, and the result of 
the Canny edge detector is quite robust to changes in lighting 
conditions.

A probabilistic Hough line detector \cite{matas2000robust} is used to extract line segments of a 
certain minimum size from the detected edges. The minimum size criterion helps 
to reject edges from white objects in the image that are not lines. The output 
line segments are filtered in the next stage to avoid false positive line 
detections where possible. We verify 
that the detected lines cover white pixels in the image, have green pixels on 
either side, and are close on both sides to edges returned by the edge detector. 
The last of these checks is motivated by the expectation that white lines, in an 
ideal scenario, will produce a pair of high responses in the edge detector, one 
on each side of the line. Ten equally spaced points are chosen on each line 
segment under review, and two normals to the line are constructed at each of 
these points, of approximate \SI{5}{\centi\metre} length in each of the two 
directions. The pixels in the captured image underneath these normals are 
checked for white colour and green colour, and the output of the canny edge 
detector is checked for a high response. The number of instances where these 
three checks succeed are independently totalled, and if all three counts exceed 
the configured thresholds, the line segment is accepted, otherwise the line 
segment is rejected.

In the final stage, similar line segments are merged together to produce fewer 
and bigger lines, as well as to cover those field line segments that might be 
partially occluded by another robot. The final result is a set of line segments 
that relate to the field lines and centre circle. Line segments that are under a 
certain threshold in length undergo a simple circle detection routine, to find 
the location of the centre circle. In our experiments, we found that this 
approach can detect circle and line segments up to \SI{4.5}{\metre} away.

\subsubsection{Localisation on the Soccer Field:}

Localisation of the robot on the soccer field---the task of estimating the 3D 
pose $(x, y, \theta)$ of the robot---is performed using the field line, centre 
circle and goal post detections. Each component of the 3D pose is estimated 
independently. To estimate the $\theta$ component, we use the global heading 
information from the magnetometer, and maintain an internal correction term 
based on the angular deviation between the expected and detected orientations of 
the white lines. This approach does not rely on having an accurate magnetometer 
output, and in experiments was able to correct deviations up to 30\degree coming 
from the magnetometer. Using the estimated $\theta$, which is normally quite 
exact, we can rotate every vision detection to align with the global field 
coordinate system. The detected line segments can thereby be classified as being 
either horizontal or vertical field lines. In each cycle of the localisation node, we use the perception information and 
dead-reckoning walking data to update the previously estimated 2D location. 
For updating 2D location, we distinguish $x$ and $y$ component using estimated $\theta$. The $y$ component of the localisation 
is updated based on the $y$ components of the detected centre circle, goal posts 
and vertical field lines. With the assumption that the robot is always inside 
the field lines, the vertical sidelines can easily be differentiated and used 
for updates. The $x$ component of the localisation is analogously updated based 
on the $x$ components of the detected centre circle, goal posts and horizontal 
field lines. The horizontal lines belonging to the goal area are discriminated 
from the centre line by checking for the presence of a consistent goal post 
detection, centre circle detection, and/or further horizontal line that is close 
and parallel. This 
approach can easily deal with common localisation difficulties, such as sensor 
aliasing and robot kidnapping. In contrast to some other proposed localisation 
methods for soccer fields, this method is relatively easy to implement and very 
robust. Our experiments indicate that the mean error of our localisation is 
better than what was reported in both \cite{laue2009efficient} and 
\cite{schulz2012utilizing}.

\subsection{Bipedal Walking}

Motivated by the changed game environment at the RoboCup competition---the 
chosen application domain for our own use of the \iguhop---the gait generation 
has been adapted to address the new challenge of walking on artificial grass. 
The use of a soft, deformable and unpredictable walking surface imposes extra 
requirements on the walking algorithm. Removable rubber cleats have been added 
at the four corners underneath each foot of the robot to improve the grip on the 
artificial grass. This also has the effect that the ground reaction forces are 
concentrated over a smaller surface area, mitigating at least part of the 
contact variability induced by the grass.

The gait is formulated in three different pose spaces: joint space, abstract 
space, and inverse space. The \emph{joint space} simply specifies all of the 
joint angles, while the \emph{inverse space} specifies the Cartesian coordinates 
and quaternion orientations of each of the limb end effectors relative to the 
trunk link frame. The \emph{abstract space} is a representation that 
was specifically developed for humanoid robots in the context of walking and 
balancing \cite{Behnke2006}. The abstract space reduces the expression of the 
pose of each limb to parameters that define the length of the limb, the 
orientation of a so-called limb centre line, and the orientation of the end 
effector. Simple conversions between all three pose spaces exist.

\begin{figure*}[!t]
\vspace{-2ex}
\parbox{\linewidth}{\centering
\includegraphics[height=25.2mm]{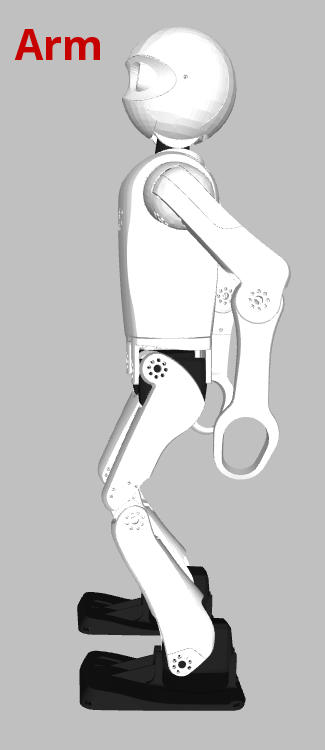}%
\includegraphics[height=25.2mm]{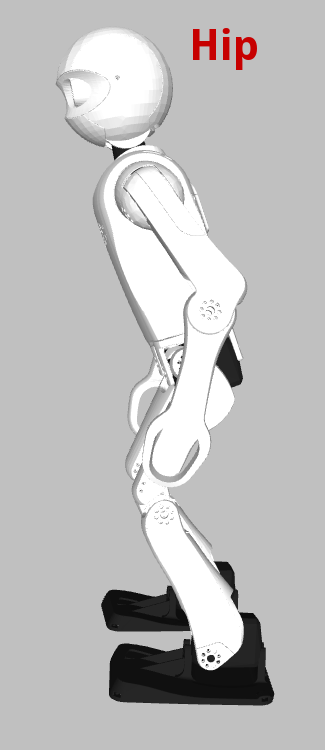}%
\includegraphics[height=25.2mm]{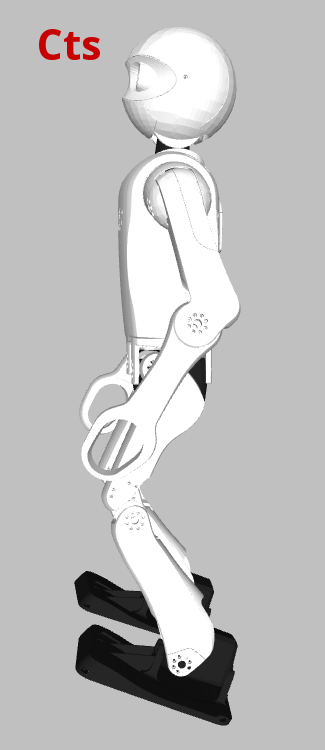}%
\includegraphics[height=25.2mm]{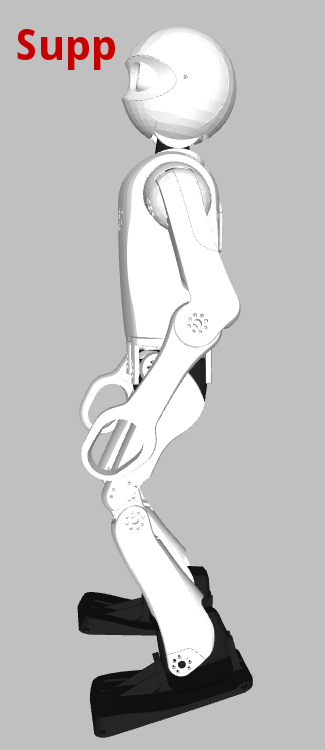}%
\includegraphics[height=25.2mm]{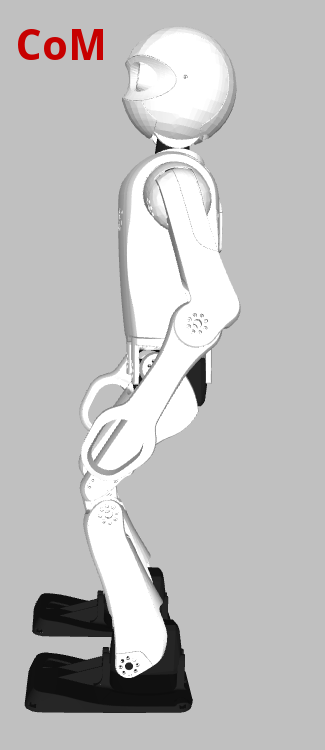}\hspace{2pt}%
\includegraphics[height=25.2mm]{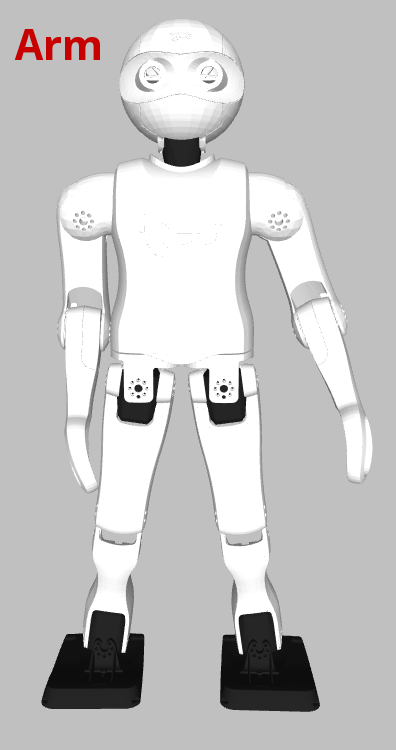}%
\includegraphics[height=25.2mm]{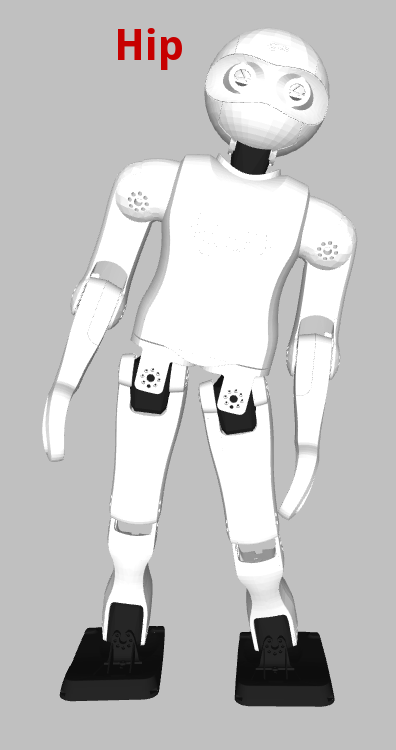}%
\includegraphics[height=25.2mm]{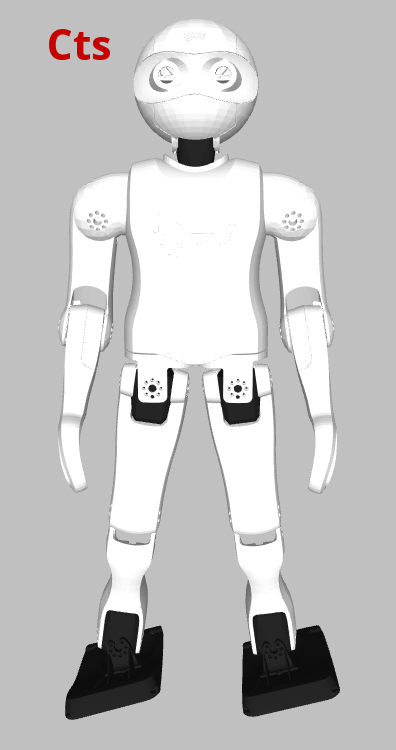}%
\includegraphics[height=25.2mm]{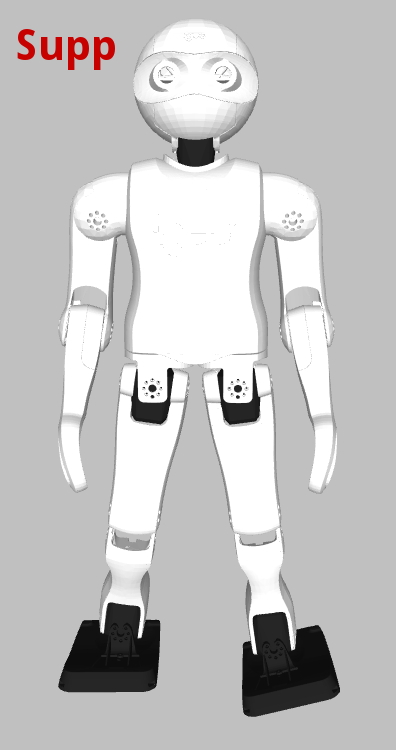}%
\includegraphics[height=25.2mm]{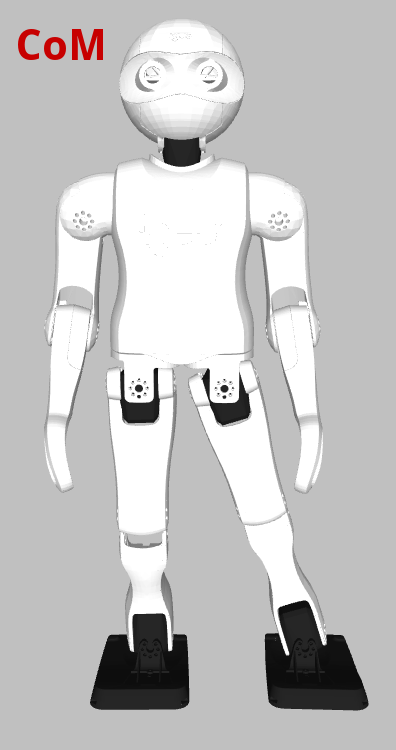}}
\caption{The implemented corrective actions in both the sagittal (left image) 
and lateral (right image) planes, from left to right in both cases the arm 
angle, hip angle, continuous foot angle, support foot angle, and CoM shifting 
corrective actions. The actions have been exaggerated for clearer illustration.}
\figlabel{corrective_actions}
\vspace{-4ex}
\end{figure*}

The walking gait in the ROS software is based on an open loop central pattern 
generated core that is calculated from a gait phase angle that increments at a 
rate proportional to the desired gait frequency. This open loop gait extends the 
gait of our previous work \cite{Missura2013a}. The central pattern generated 
gait begins with a configured halt pose in the abstract space, then incorporates 
numerous additive waveforms to the halt pose as functions of the gait phase and 
commanded gait velocity. These waveforms generate features such as leg lifting, 
leg swinging, arm swinging, and so on. The resulting abstract pose is converted 
to the inverse space, where further motion components are added. The resulting 
inverse pose is converted to the joint space, in which form it is commanded to 
the robot  actuators. A pose blending scheme towards the 
halt pose is implemented in the final joint space representation to smoothen the 
transitions to and from walking.

A number of simultaneously operating basic feedback mechanisms have been built 
around the open loop gait core to stabilise the walking. The PID-like feedback 
in each of these mechanisms derives from the fused pitch and fused roll 
\cite{Allgeuer2015} state estimates and works by adding extra corrective action 
components to the central pattern generated waveforms in both the abstract and 
inverse spaces, namely arm angle, hip angle, continuous foot angle, support foot 
angle, CoM shifting, and virtual slope. The corrective actions are illustrated in 
\figref{corrective_actions}. The step timing is computed using the capture step 
framework \cite{missura2014balanced}, based on the lateral CoM state \cite{Allgeuer2016a}.

Overall, the feedback mechanisms were observed to make a significant difference 
in the walking ability of the robots, with walking often not even being possible 
for extended periods of time without them. The feedback mechanisms also imparted 
the robots with disturbance rejection capabilities that were not present 
otherwise. Reliable omnidirectional walking speeds of 
\SI{21}{\centi\metre\per\second} were achieved on an artificial grass surface of 
blade length \SI{32}{\milli\metre}. Over all games played at RoboCup 2016, none of 
the five robots ever fell while walking\footnote{Video: 
\url{https://www.youtube.com/watch?v=9saVpA3wIbU}} in free space. Only strong 
collisions with other robots caused falls from walking Igus robot s could quickly recover using 
keyframe get-up motions \cite{stuckler2006getting}.

\subsection{Soccer Behaviours}

Given the current game state, as perceived by the vision system, the robots must 
autonomously decide on the higher-level actions to execute in order to try to 
score a goal. For this we use a two-layered hierarchical finite state machine 
(FSM), that has a tailor-made custom implementation for RoboCup soccer. The 
upper layer is referred to as the \emph{game FSM}, and the lower layer is 
referred to as the \emph{behaviour FSM}. Given the current required playing 
state of the robot, the former is responsible for deciding on a suitable higher-level
action, such as for example ``dribble or kick the ball to these specified 
target coordinates'', and based on this higher-level action, the latter is 
responsible for deciding on the required gait velocity, whether to kick or dive, 
and so on.

In the order of execution, a ROS interface module first abstracts away the 
acquisition of data into the behaviours node, before this data is then 
processed, refined and accumulated into a so-called sensor variables structure. 
This aggregates all kinds of information from the vision, localisation, RoboCup 
game controller, team communications and robot control nodes, and precomputes 
commonly required derived data, such as for example the coordinates of the 
currently most intrusive obstacle, whether the current ball estimate is stable, 
and/or how long ago it was last seen. This information is used to decide on the 
appropriate game FSM state, such as for example \emph{default ball handling}, 
\emph{positioning}, or \emph{wait for ball in play}, which is then executed and 
used to compute standardised game variables, such as for example \emph{kick if 
possible} and \emph{ball target}. These game variables, along with the sensor 
variables, are then used by the behaviour FSM to decide on a suitable state, 
such as for example \emph{dribble ball}, \emph{walk to pose} or \emph{go behind 
ball}. The execution of the required behaviour state then yields the required 
low-level action of the robot, which is passed to the robot control node via the 
aforementioned ROS interface module, completing the execution of the soccer 
behaviours.

\subsection{Human-Robot Interfaces}

Despite being designed to operate autonomously, our robots still need 
suitable human-robot interfaces to allow them to be configured and calibrated. For the lowest and 
most fundamental level of control and operation, each robot can be launched and 
configured directly on the command line inside SSH sessions directly on the 
robot PC. This allows the greatest amount of freedom and flexibility in 
launching ROS nodes and checking their correct operation, but is also a complex 
and time-consuming task that is prone to errors and requires in-depth 
knowledge of the robot and software framework.

\begin{figure}[!ht]
\vspace{-1ex}
\parbox{\linewidth}{\centering\includegraphics[width=0.6\linewidth]{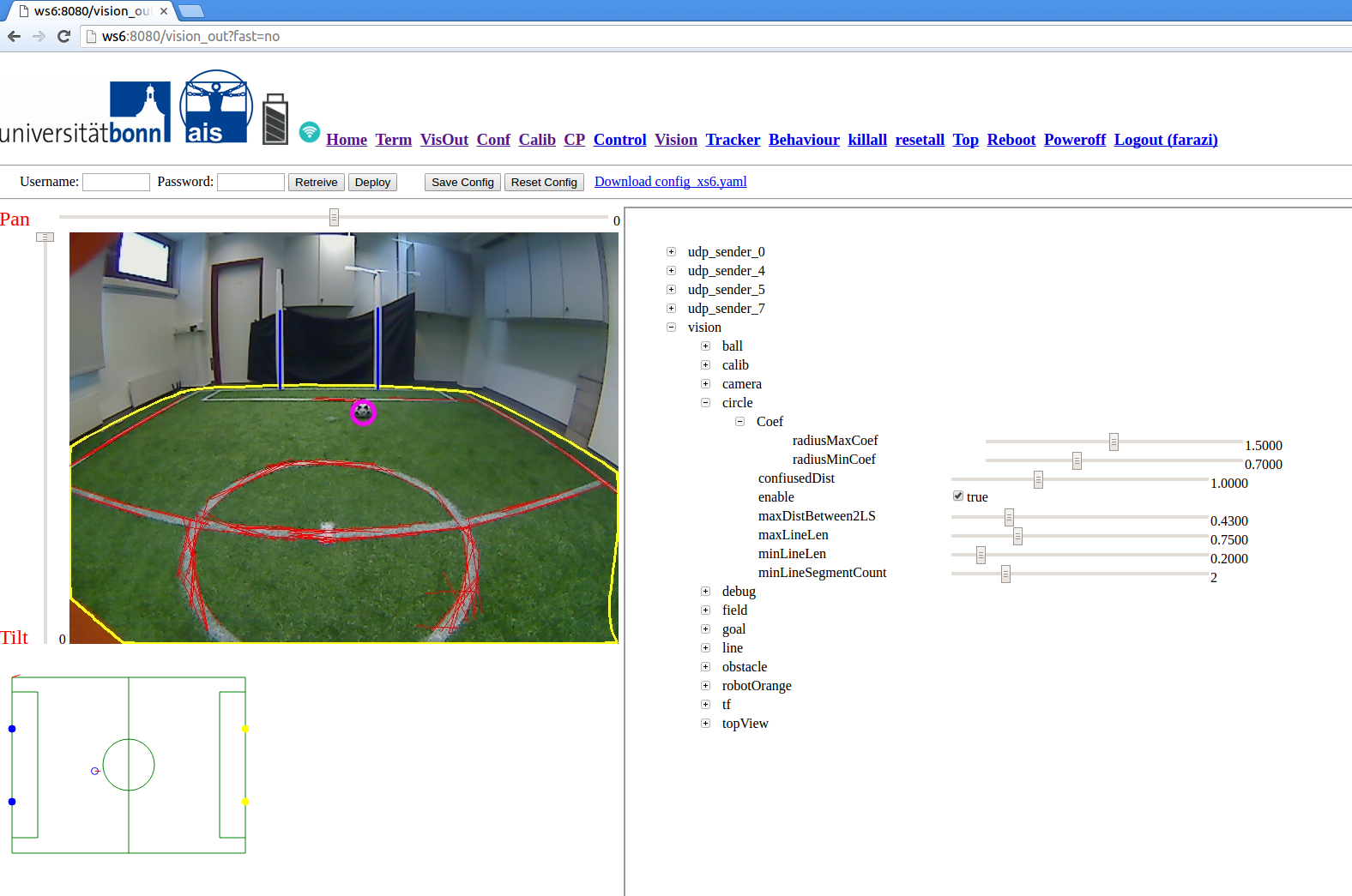}}
\caption{A screenshot of the web application used to help calibrate the robot.}
\figlabel{web_server}
\vspace{-4ex}
\end{figure} 

To overcome these problems, a web application was developed for the robot (see 
\figref{web_server}), with the robot PC as the web server, to allow standard web 
browsers of all devices to connect to the robot for configuration and calibration. This operates 
at a higher level of abstraction than the command line, and allows users to 
perform all common tasks that are required when operating the robot. This makes 
routine tasks significantly quicker and easier than on the command line, and 
avoids problems altogether such as hangup signals and resuming command line 
sessions. By exploiting the client-server architecture of web applications and 
the highly developed underlying web protocols, the connection is very robust, 
even over poor quality wireless network connections, and of low computational 
cost for the robot as most processing is implemented on the client side. The web 
application, amongst many other things, allows the user to start, stop and 
monitor ROS nodes, displays all kinds of status information about the robot, 
allows dynamic updates to the configuration server parameters, shows the 
processed vision and localisation outputs, allows various calibration and system 
services to be called, and allows the pose of the head to be controlled 
manually.

During operation, whether managed over the command line or the web server, the 
robot can be visualised using the RQT GUI, and dynamically reconfigured using 
the configuration server. This requires a live network connection to the robot 
for communication purposes. To configure the robot instantaneously, and without 
the need for any kind of network connection, a QR code detector has been 
implemented in the vision module. With this feature, arbitrary reconfiguration 
tasks can be effectuated due to the great freedom of data that can be robustly 
encoded in a QR code. QR codes can conveniently be generated on mobile devices, 
also for example with a dedicated mobile application, and shown to the robot at 
any time. The robot plays a short tune to acknowledge the QR code, serving as 
auditory feedback that the QR code was detected and processed.

\section{Conclusions}

In this paper, we described our platforms and approaches to playing soccer in 
the Humanoid TeenSize class. During RoboCup 2016, we successfully demonstrated that our 
robots could robustly perceive the game environment, make decisions, and act on 
them. Our team NimbRo TeenSize aggregated a total score of 29:0 over five games. 
We have released our hardware\footnote{Hardware: 
\url{https://github.com/igusGmbH/HumanoidOpenPlatform}} and 
software\footnote{Software: \url{https://github.com/AIS-Bonn/humanoid_op_ros}} 
to GitHub with the hope that it is beneficial for other teams and research 
groups.

\section{Acknowledgements}
\footnotesize
We acknowledge the contributions of \igus GmbH to the project, in particular the 
management of Martin Raak towards the robot design and manufacture. This work 
was partially funded by grant BE 2556/10 of the German Research Foundation 
(DFG).

\bibliographystyle{ieeetr}
\bibliography{ms}

\end{document}